# The Role of Artificial Intelligence in Enhancing Insulin Recommendations and Therapy Outcomes

Maria Panagiotou[1], Knut Strømmen[1], Lorenzo Brigato[1], Bastiaan E. de Galan[2,3,4], and Stavroula Mougiakakou[1]

[1] ARTORG Center of Biomedical Engineering Research, University of Bern, Switzerland
[2]CARIM School for Cardiovascular Diseases, Maastricht University, Maastricht, The Netherlands.
[3]Department of Internal Medicine, Maastricht University Medical Centre+, Maastricht, The Netherlands.
[4]Department of Internal Medicine, Radboud University Medical Centre, Nijmegen, The Netherlands
Correspondence address: stavroula.mougiakakou@unibe.ch

**Abstract.** The growing worldwide incidence of diabetes requires more effective approaches for managing blood glucose levels. Insulin delivery systems have advanced significantly, with artificial intelligence (AI) playing a key role in improving their precision and adaptability. AI algorithms, particularly those based on reinforcement learning, allow for personalised insulin dosing by continuously adapting to an individual's responses. Despite these advancements, challenges such as data privacy, algorithm transparency, and accessibility still need to be addressed. Continued progress and validation in AI-driven insulin delivery systems promise to improve therapy outcomes further, offering people more effective and individualised management of their diabetes. This paper presents an overview of current strategies, key challenges, and future directions.

## 1 Introduction

The escalating global burden of people living with diabetes underscores the need for innovative approaches to optimise insulin management strategies for those requiring insulin. Unlike many other medications, insulin needs are highly variable, which means that the required dose need to be titrated on the basis of real-time indicators such as current blood glucose levels, the carbohydrate content of meals and the level of physical activity, to name a few of the dozens of factors involved. This process makes insulin dosing profoundly challenging and requires people to perform complex calculations to determine the best dose. Non-optimal insulin dosing can lead to serious adverse effects, particularly hypo- or hyperglycaemia, putting people at risk of vascular complications. Integrating artificial intelligence (AI) into insulin recommendation systems represents a significant advancement, enabling precise and adaptive insulin dosing, which can enhance treatment outcomes. In this paper, we provide an overview of existing strategies, challenges, and future prospects.

# 2 Insulin Recommendation Systems

## 2.1 Evolution of Insulin Recommendation Systems

Devices commonly used for insulin delivery include insulin pens and pumps. Innovations in continuous glucose monitoring (CGM) and AI algorithms have significantly advanced the development of these systems.

Fundamental to these advancements are concepts such as hybrid closed-loop, fully closed-loop control [1], and open-loop control. Fully closed-loop control, often referred to as an artificial pancreas (AP), uses a control algorithm as its "brain" to analyse CGM data and automatically adjust insulin infusion rates, ensuring blood glucose levels stay within a target range [1], [2]. In contrast, open-loop control relies on manual insulin adjustments based on external input factors such as meals and activity [3]. Hybrid closed-loop systems combine elements of both closed-loop and open-loop control, typically allowing for automated insulin delivery, but the user is still required to manually program insulin boluses with meals [1].

Historically, algorithmic approaches for insulin recommendation have centred around bolus calculators [4], model predictive control [5], [6], proportional-integral-derivative, fuzzy logic [7], learning algorithms [8], and Kalman filters [9]. Up to now, randomised clinical trials on bolus advisors in people with diabetes have shown modest improvements in HbA1c or other glycaemic outcomes and treatment satisfaction [10]. One of the earliest works using AI for insulin adjustment was introduced in 1998 and published a few years later. Specifically, Mougiakakou et al. [11] developed a neural network-based decision support system to guide insulin regimen (six different insulin regimes, as a combination of short- and intermediate-acting insulin) and dose adjustments for individuals with type 1 diabetes under multiple daily injection (MDI) and utilising self-monitoring blood glucose (SMBG) measurements. Over the years, the increasing availability of data and smartphones has facilitated the introduction of numerous data-driven approaches aimed at optimising insulin dosing [10] Tyler et al. [11] utilised k-nearest-neighbour methods to generate recommendations for optimal insulin dosing in the context of a quality control algorithm for people with type 1 diabetes (T1D) under MDI therapy. Pesl et al. [12] used case-based reasoning in the advanced bolus calculator for diabetes to provide meal-time dosing advice.

## 2.2 Reinforcement Learning-based Algorithms for Insulin Recommendation Systems

To further improve the self-management of people with diabetes, automatically personalising adjustment of insulin intake in the field is essential. One such approach that has gained traction in insulin delivery is reinforcement learning (RL), which offers automatic continuous adaptation and exploration of better solutions [14].

RL is a machine learning framework for learning sequential decision-making tasks. It is designed for problems involving a learning agent interacting with its environment to achieve a goal. In insulin delivery systems for diabetes, blood glucose control is achieved through a learning agent, acting as the controller, interacting with the environment—which is the individual's body [13]. The ability to continuously learn and adapt to individual responses and improve decision-making in highly complex environments position this technology as a potential breakthrough in disease self-management and the personalisation of treatment strategies. Within this context, two primary approaches are used in RL: model-based and model-free RL. Model-based approaches require a predefined model describing the relationship between insulin dosing and glucose response, while model-free approaches learn directly from observed data without a predefined physiological model.

Recent advancements in RL-driven insulin delivery are evident in both closed-loop and open-loop systems. These approaches leverage RL algorithms to optimise glucose control, leading to improved therapy outcomes. Table 1 provides an overview of relevant literature on RL applications in insulin delivery systems, and the following sections will explore key developments in this field.

**Closed/hybrid-loop automated insulin delivery system**

Several RL methods have emerged to tailor insulin administration for people with T1D in the past few years. Pioneering work by Daskalaki et al. [14], [15], [16] proposed an actor-critic (AC) method initialised with information transferred from insulin to glucose signals. Sun et al. [17] extended and further validated this algorithm. More recently, studies have increasingly integrated CGMs into advanced algorithms which include neural networks for managing T1D. For instance, a modular deep RL algorithm based on the proximal policy optimisation algorithm was designed to fully automate glucose control, utilising CGMs [18]. Similarly, Zhu et al. [19] developed a deep-Q learning agent (DQN) to suggest basal insulin values and a deep deterministic policy gradient AC model for insulin bolus control, both supported by CGMs [20]. Another study by Jafar et al. [21] proposed a Q-learning approach to adaptively optimise carbohydrate ratios and basal rates, leveraging CGMs for continuous feedback. Fox et al. [22] employed the soft AC algorithm to develop glucose control policies for closed-loop blood glucose control with CGMs. Furthermore, recent work combined evolutionary, DQN, AC, and uncertainty estimation algorithms to adjust insulin sensitivity and carbohydrate-to-insulin ratios for meal boluses and reference basal rates in pump therapy and insulin pen usage, all facilitated by continuous data from CGMs [23].

Previous work has rarely applied RL to type 2 diabetes (T2D). A recent study proposed a model-based RL method to determine the optimal insulin regimen by evaluating rewards associated with the glycaemic state through interactions with a patient model that models the individual glycaemic responses to insulin dosages. The algorithm outperformed other methods in insulin titration optimisation and showed promising results in a blinded-feasibility trial [24].

**Open-loop insulin delivery system**

Sun et al. [40] extended and further validated an AC model-free approach initially introduced for CGM and pump therapy to SMBG measurements and MDI therapy. A similar approach proposed a complete AI-based system incorporating an algorithm for carbohydrate estimation based on food images and an AC method to recommend basal and bolus insulin [37]. A recent study proposed an offline RL approach with online fine-tuning for a dual basal-bolus calculator for people with T1D under CGM monitoring and MDI treatment [39].

Liu et al. [33] developed an algorithm based on the DQN to recommend the number of oral antidiabetic drugs and insulins for people with T2D. The evaluation of the algorithm involved assessing its prescriptions' concordance with recommendations and comparing clinical outcomes, showing significantly better long-term outcomes and reduced hypoglycaemia events compared to traditional methods.

## 3 Factors Considered in Enhancing Therapy Outcomes

Numerous factors are crucial for improving therapy outcomes in insulin delivery systems. Key considerations highlighted in recent RL literature include:

- **Personalised model-free online learning approaches**: Real-time personalisation using model-free online learning approaches allows for more accurate and adaptive insulin dosing without relying on pre-existing models, enhancing individual care. Most authors based their algorithms on model-free methods as they provide a personalised solution that learns directly from the person. This approach accounts for the inter- and intra-subject variability of insulin absorption and insulin action, which differs from person to person. Simulation results in the literature show the effectiveness of these algorithms in improving glucose control (14–17,21,26,27,32, 39).
- **Closed-loop algorithms without meal announcement and carbohydrate estimation**: Developing closed-loop systems that function effectively without requiring meal announcements or precise carbohydrate estimation represents a major advancement in the field. As an algorithm becomes more sophisticated and capable of compensating for uncertainties, it reduces the dependency on precise meal and portion size announcements. While current results in the literature highlight their potential, improvements are still needed to optimise these systems and achieve the desired outcomes [18], [22], [36].

| **Paper** | **Year** | **Population** | | | **Devices** | | **Algorithm** | **Adaptation** | **Studies** | |
|---|---|---|---|---|---|---|---|---|---|---|
| | | *Adults* | *Adolescents* | *Children* | *Input* | *Output* | | | *In-silico* | *Clinical* |
| Daskalaki E. et al. **[14]** | 2013 | 10, T1D | 10, T1D | 8, T1D | CGM | Pump | AC | BR<br>ICR | UVA/PadovaT1D [25]<br>Duration: 30 days | |
| Daskalaki E. et al. **[15]** | 2013 | 10, T1D | 10, T1D | 8, T1D | CGM | Pump | AC | BR<br>ICR | UVA/Padova T1D [25]<br>Duration: 10 days | |
| Paula M. et al. **[26]** | 2015 | NA | NA | NA | CGM | Pump | On-line policy learning | Insulin Infusion rate | Custom<br>Duration: 1 day | |
| Daskalaki E. et al. **[16]** | 2016 | 110, T1D | 10, T1D | 8, T1D | CGM | Pump | AC | BR<br>ICR | UVA/Padova T1D [25]<br>Duration: 14 days | |
| Sun Q. et al. **[17]** | 2018 | 100, T1D | 0 | 0 | CGM<br>SMBG | Pump | AC | BR<br>ICR | UVA/Padova T1D [25]<br>Duration: 90 days | |
| Sun Q. et al. **[27]** | 2019 | 10, T1D | 0 | 0 | SMBG | MDI | AC | Basal insulin<br>ICR | DMMS.R [28]<br>Duration: 15 days | |
| Zhu T. et al. **[19]** | 2019 | 10, T1D | 10, T1D | 0 | CGM | Pump | DDQN | Basal insulin<br>Glucagon | UVA/Padova T1D [25]<br>Duration: 180 days | |
| Zhu T. et al. **[20]** | 2020 | 10, T1D | 10, T1D | 0 | CGM | Pump<br>MDI | DDPG | Bolus insulin | UVA/Padova T1D [25]<br>Duration: 90 days | |
| Fox I. et al. **[22]** | 2020 | 10, T1D | 10, T1D | 10, T1D | CGM | Pump | SAC | BR | simglucose [29]<br>Duration: 10 days | |
| Zhu T. et al. **[30]** | 2021 | 10, T1D | 10, T1D | 0 | CGM | Pump | DQN | BR | UVA/Padova T1D [25]<br>Duration: 90 days | |
| Jafar A. et al. **[21]** | 2021 | 50, T1D | 0 | 0 | CGM | Pump | Q-learning | BR, ICR | Hovorka's model [31]<br>Duration: 35 days | |
| Krishnamoorthy D. et al. **[32]** | 2020 | 50, T2D | 0 | 0 | SMBG | MDI | Recursive least square-based extremum seeking control. | Basal insulin | Custom<br>Duration: 60 days | |
| Liu et al. **[33]** | 2020 | 2,847, T2D | 0 | 0 | NA | NA | DQN | Number of OAD and Insulins | NEW2D dataset [34]<br>Duration: 5 visits | |
| Hettiarachchi C. et al. **[18]** | 2022 | 10, T1D | 0 | 0 | CGM | Pump | PPO | Insulin Infusion rate | simglucose [29]<br>Duration: 26 hours | |
| Louis M. et al. **[23]** | 2022 | 10, T1D | 10, T1D | 0 | CGM | Pump | Evolution strategies | Insulin Infusion rate | UVA/Padova T1D [25]<br>Duration: 10 days | |
| Emmerson H. et al. **[35]** | 2023 | 10, T1D | 10, T1D | 10, T1D | CGM | Pump | Offline RL | BR | simglucose [29]<br>Duration: 10 days | |
| Giorno S. et al. **[36]** | 2023 | 10, T1D | 0 | 0 | CGM | Pump | DQN | Bolus insulin | UVA/Padova T1D [25]<br>Duration: 6 to 24 hours | |
| Wang G. et al. **[24]** | 2023 | 16, T2D | 0 | 0 | CGM | MDI | Model-based | Basal-bolus insulin | | Retrospective comparison of system's vs HCP's suggestions.<br>Duration: 7 days |
| Panagiotou M. et al. **[37]** | 2023 | 11, T1D | 0 | 0 | SMBG | MDI | AC | Basal Insulin, ICR | DMMS.R [28]<br>Duration: 14 days | |
| Jafar A. et al. **[38]** | 2024 | 100, T1D (in-silico)<br>4, T1D<br>19, T2D | 0 | 0 | CGM | MDI | Q-learning | ICR, CF | UVA/Padova T1D [25]<br>Duration: 60 days | Retrospective comparison of system's vs HCP's suggestions.<br>Duration:16 weeks |
| Yoo J. et al. **[39]** | 2024 | 10, T1D | 10, T1D | 10, T1D | CGM | MDI | Offline RL | Basal insulin<br>ICR | simglucose [29]<br>Duration: 60 days | |

**Table 1.** Literature Overview of RL Approaches for Insulin Recommendation. T1D: Type 1 Diabetes, T2D: Type 2 Diabetes, CGM: Continues Glucose Monitoring, SMBG: Self-Monitoring Blood Glucose, MDI: Multiple Daily Injections, AC: Actor-Critic, DDQN: Double Deep Q Networks, DDPG: Deep Deterministic Policy Gradient, DQN: Deep Q Networks, SAC: Soft Actor-Critic, PPO: Proximal Policy Optimisation, BR: Basal Rate, ICR: Insulin-to-Carbohydrate Ratio, OAD: Oral antidiabetic drug, NEW2D: Newly Diagnosed Type 2 Diabetes Patients in China.

- **AI-based dietary assessment and carbohydrate estimation**: AI-powered systems for more accurate dietary assessments and carbohydrate estimations significantly enhance therapy outcomes by providing more accurate insulin dose recommendations compared to traditional manual carbohydrate estimation and standard bolus calculators [37]. This approach is particularly beneficial for hybrid- or open-loop systems, as it minimises user input, reducing the likelihood of errors, and making the system more autonomous.
- **Insulin-on-board and insulin sensitivity factors**: Incorporating insulin-on-board calculations and factoring in insulin sensitivity variations help optimise insulin dosing. Studies have demonstrated that considering these factors significantly improves the percentage of time spent in the target glucose range while reducing hypoglycemia compared to baseline methods in in-silico settings. Additionally, these factors have been evaluated by comparing the algorithm recommendations with the endocrinologist's recommendations and real-world individual therapy settings (20,24).

  **Personalised insulin dosing for high-fat meals and postprandial aerobic exercise**: Research indicates that tailoring insulin doses based on the specific composition of high-fat meals and postprandial exercise improves postprandial glucose outcomes and reduces time in hypoglycemia after high-fat meals and postprandial exercises [41].

# 4 Challenges and Mitigations

Several challenges must be addressed to ensure a broader acceptance and efficacy in RL-based insulin delivery systems. Key challenges and their potential mitigations are as follows:

- **Algorithm Evaluation:** Most of the algorithms in the literature used in-silico simulators and compared the results with a baseline method, often a standard meal bolus calculator and a standard basal insulin dose or a low-glucose insulin suspension (LGS) strategy. They all show superior results regarding time spent in the target range, time above range, and time below range. Nevertheless, the lack of standardised meal protocol and simulator benchmarks complicates consistent evaluation and comparative analysis across interventions and diabetes groups. Despite promising results in in-silico environments, the clinical validation of AI-based approaches remains limited, with only a few having undergone trials in real-world settings [42], [43]. Therefore, it is essential to conduct more clinical trials to confirm their effectiveness and ensure their reliable performance in diverse patient populations and varying conditions.
- **Algorithm Transparency:** The complexity of RL algorithms often raises concerns for users regarding their acceptance of the insulin suggestion. To secure individuals' and clinicians' trust, we need to ensure the explainability and interpretability of the systems. Developing user-friendly interfaces and providing straightforward explanations of the algorithm's decisions can help bridge this gap. Transparent communication about algorithm functionality can further empower users to accept treatment recommendations.

- **Access and Affordability:** A significant barrier to adopting AI-driven insulin delivery systems is the reliance on closed-loop frameworks, which require expensive devices such as CGMs and insulin pumps. These technologies are often beyond the financial reach of many people. To address this, researchers have developed cost-effective alternatives using traditional devices, such as SMBG meters and insulin pens, where RL algorithms are applied and make advanced diabetes management more accessible to a broader population [17], [37], [39]. These solutions not only democratise access to healthcare services but are also agnostic to specific diabetes treatment devices, offering the potential to be used by individuals in underserved areas with limited access to healthcare systems.
- **People Engagement:** Diabetes self-management depends heavily on regular monitoring and documenting factors such as blood glucose levels, insulin doses, and dietary habits. However, this can be particularly overwhelming for older people or those with low health literacy. To tackle this, tools should be developed to assist people in managing their condition without adding burden. For example, automated data collection, assistance on dietary assessments, reminders, and tracking apps can enhance engagement while simplifying self-management.
- **Data Privacy and Security:** Insulin delivery systems require extensive data collection, including sensitive health information, to propose a precise insulin recommendation. Ensuring compliance with data protection regulations (such as GDPR or HIPAA), employing robust encryption methods, and implementing secure data storage protocols are essential for safeguarding individual information. Fostering trust through transparent data policies and anonymised data usage can help address privacy concerns.
- **Regulatory Framework:** The regulatory framework for AI in medicine presents an important consideration. While AI-driven medical solutions hold great promise, their deployment must comply with evolving regulations such as the EU AI Act, FDA guidelines, and MDR (Medical Device Regulation) to ensure safety, efficacy, and accountability. The regulatory landscape must adapt to the unique nature of AI technologies, which continuously evolve and learn. Furthermore, translating ethical principles, such as fairness, transparency, and patient autonomy, into practical applications remains a significant challenge. Ensuring that AI systems align with these ethical guidelines in diverse clinical settings requires continuous oversight, interdisciplinary collaboration, and adaptive governance frameworks to ensure that patient care is not compromised and that AI technologies are used responsibly.

## 5 Conclusion

Integrating AI into insulin delivery systems represents a transformative advancement in diabetes management, offering real-time personalisation, improved glycaemic control, and better outcomes. Notably, advancements in algorithms, particularly in RL, have enabled systems to adapt to individual variability and complex dynamic factors such as insulin absorption rates and glucose response patterns. However, to further improve the effectiveness of these systems, future research must address additional critical dynamic factors, including insulin sensitivity, physical activity, sleep quality, illness, and the impact of other macronutrients like protein and fat [44]. Achieving this will require the development of advanced simulators or datasets capable of replicating the complex interactions of these variables. To fully realise the potential of these technologies, further research and closer collaboration between engineers, computer scientists in AI, health care professionals, and people with diabetes are essential to validate and optimise AI-powered insulin delivery systems in real-world clinical settings across different setups.

## Funding

This research has been funded by the European Commission and the Swiss Confederation-State Secretariat for Education, Research and Innovation (SERI) within the project 101057730 Mobile Artificial Intelligence Solution for Diabetes Adaptive Care (MELISSA).

## References

[1] S. Templer, 'Closed-Loop Insulin Delivery Systems: Past, Present, and Future Directions', *Front Endocrinol (Lausanne)*, vol. 13, p. 919942, 2022, doi: 10.3389/fendo.2022.919942.

[2] C. Cobelli, E. Renard, and B. Kovatchev, 'Artificial pancreas: past, present, future', *Diabetes*, vol. 60, no. 11, pp. 2672–2682, Nov. 2011, doi: 10.2337/db11-0654.

[3] D. Lian, 'Insulin delivery devices for diabetes mellitus from open-loop to closed-loop', *AIP Conference Proceedings*, vol. 2350, no. 1, p. 020002, Apr. 2021, doi: 10.1063/5.0049209.

[4] R. Ziegler *et al.*, 'Use of an insulin bolus advisor improves glycemic control in multiple daily insulin injection (MDI) therapy patients with suboptimal glycemic control: first results from the ABACUS trial', *Diabetes Care*, vol. 36, no. 11, pp. 3613–3619, Nov. 2013, doi: 10.2337/dc13-0251.

[5] T. B. Aradóttir *et al.*, 'Model predictive control for dose guidance in long acting insulin treatment of type 2 diabetes', *IFAC Journal of Systems and Control*, vol. 9, p. 100067, Sep. 2019, doi: 10.1016/j.ifacsc.2019.100067.

[6] S. G. Mougiakakou *et al.*, 'SMARTDIAB: a communication and information technology approach for the intelligent monitoring, management and follow-up of type 1 diabetes

patients', *IEEE Trans Inf Technol Biomed*, vol. 14, no. 3, pp. 622–633, May 2010, doi: 10.1109/TITB.2009.2039711.

[7] E. Atlas, R. Nimri, S. Miller, E. A. Grunberg, and M. Phillip, 'MD-logic artificial pancreas system: a pilot study in adults with type 1 diabetes', *Diabetes Care*, vol. 33, no. 5, pp. 1072–1076, May 2010, doi: 10.2337/dc09-1830.

[8] M. Rigla, G. García-Sáez, B. Pons, and M. E. Hernando, 'Artificial Intelligence Methodologies and Their Application to Diabetes', *J Diabetes Sci Technol*, vol. 12, no. 2, pp. 303–310, Mar. 2018, doi: 10.1177/1932296817710475.

[9] S. E. Engell, T. B. Aradóttir, T. K. S. Ritschel, H. Bengtsson, and J. B. Jørgensen, 'Estimating a Personalized Basal Insulin Dose from Short-Term Closed-Loop Data in Type 2 Diabetes', in *2022 IEEE 61st Conference on Decision and Control (CDC)*, Dec. 2022, pp. 2580–2585. doi: 10.1109/CDC51059.2022.9992960.

[10] E. J. den Brok *et al.*, 'The effect of bolus advisors on glycaemic parameters in adults with diabetes on intensive insulin therapy: A systematic review with meta-analysis', *Diabetes Obes Metab*, vol. 26, no. 5, pp. 1950–1961, May 2024, doi: 10.1111/dom.15521.

[11] N. S. Tyler *et al.*, 'An artificial intelligence decision support system for the management of type 1 diabetes', *Nat Metab*, vol. 2, no. 7, pp. 612–619, Jul. 2020, doi: 10.1038/s42255-020-0212-y.

[12] P. Pesl *et al.*, 'Case-Based Reasoning for Insulin Bolus Advice', *J Diabetes Sci Technol*, vol. 11, no. 1, pp. 37–42, Jan. 2017, doi: 10.1177/1932296816629986.

[13] M. Tejedor, A. Z. Woldaregay, and F. Godtliebsen, 'Reinforcement learning application in diabetes blood glucose control: A systematic review', *Artif Intell Med*, vol. 104, p. 101836, Apr. 2020, doi: 10.1016/j.artmed.2020.101836.

[14] E. Daskalaki, P. Diem, and S. G. Mougiakakou, 'An Actor-Critic based controller for glucose regulation in type 1 diabetes', *Comput Methods Programs Biomed*, vol. 109, no. 2, pp. 116–125, Feb. 2013, doi: 10.1016/j.cmpb.2012.03.002.

[15] E. Daskalaki, P. Diem, and S. G. Mougiakakou, 'Personalized tuning of a reinforcement learning control algorithm for glucose regulation', *Annu Int Conf IEEE Eng Med Biol Soc*, vol. 2013, pp. 3487–3490, 2013, doi: 10.1109/EMBC.2013.6610293.

[16] E. Daskalaki, P. Diem, and S. G. Mougiakakou, 'Model-Free Machine Learning in Biomedicine: Feasibility Study in Type 1 Diabetes', *PLOS ONE*, vol. 11, no. 7, p. e0158722, Jul. 2016, doi: 10.1371/journal.pone.0158722.

[17] Q. Sun *et al.*, 'A dual mode adaptive basal-bolus advisor based on reinforcement learning', *IEEE J. Biomed. Health Inform.*, vol. 23, no. 6, pp. 2633–2641, Nov. 2019, doi: 10.1109/JBHI.2018.2887067.

[18] C. Hettiarachchi, N. Malagutti, C. Nolan, E. Daskalaki, and H. Suominen, 'A Reinforcement Learning Based System for Blood Glucose Control without Carbohydrate Estimation in Type 1 Diabetes: In Silico Validation', in *2022 44th Annual International Conference of the IEEE Engineering in Medicine & Biology Society (EMBC)*, Jul. 2022, pp. 950–956. doi: 10.1109/EMBC48229.2022.9871054.

[19] T. Zhu, K. Li, and P. Georgiou, 'A Dual-Hormone Closed-Loop Delivery System for Type 1 Diabetes Using Deep Reinforcement Learning', Oct. 09, 2019, *arXiv*: arXiv:1910.04059. doi: 10.48550/arXiv.1910.04059.

[20] T. Zhu, K. Li, L. Kuang, P. Herrero, and P. Georgiou, ‘An Insulin Bolus Advisor for Type 1 Diabetes Using Deep Reinforcement Learning’, *Sensors*, vol. 20, no. 18, Art. no. 18, Jan. 2020, doi: 10.3390/s20185058.

[21] A. Jafar, A. E. Fathi, and A. Haidar, ‘Long-term use of the hybrid artificial pancreas by adjusting carbohydrate ratios and programmed basal rate: A reinforcement learning approach’, *Comput Methods Programs Biomed*, vol. 200, p. 105936, Mar. 2021, doi: 10.1016/j.cmpb.2021.105936.

[22] I. Fox, J. Lee, R. Pop-Busui, and J. Wiens, ‘Deep Reinforcement Learning for Closed-Loop Blood Glucose Control’, Sep. 18, 2020, *arXiv*: arXiv:2009.09051. Accessed: Nov. 16, 2023. [Online]. Available: http://arxiv.org/abs/2009.09051

[23] M. Louis, H. R. Ugalde, P. Gauthier, A. Adenis, Y. Tourki, and E. Huneker, ‘Safe Reinforcement Learning for Automatic Insulin Delivery in Type I Diabetes’, in *Reinforcement Learning for Real Life Workshop, NeurIPS 2022*, New Orleans (LA), United States, Dec. 2022. Accessed: Nov. 16, 2023. [Online]. Available: https://hal.science/hal-03939853

[24] G. Wang *et al.*, ‘Optimized glycemic control of type 2 diabetes with reinforcement learning: a proof-of-concept trial’, *Nat Med*, vol. 29, no. 10, pp. 2633–2642, Oct. 2023, doi: 10.1038/s41591-023-02552-9.

[25] ‘T1DMS – The Epsilon Group’. Accessed: Nov. 16, 2023. [Online]. Available: https://tegvirginia.com/software/t1dms/

[26] M. De Paula, G. G. Acosta, and E. C. Martínez, ‘On-line policy learning and adaptation for real-time personalization of an artificial pancreas’, *Expert Systems with Applications*, vol. 42, no. 4, pp. 2234–2255, Mar. 2015, doi: 10.1016/j.eswa.2014.10.038.

[27] Q. Sun, M. V. Jankovic, and S. G. Mougiakakou, ‘Reinforcement Learning-Based Adaptive Insulin Advisor for Individuals with Type 1 Diabetes Patients under Multiple Daily Injections Therapy’, *Annu Int Conf IEEE Eng Med Biol Soc*, vol. 2019, pp. 3609–3612, Jul. 2019, doi: 10.1109/EMBC.2019.8857178.

[28] ‘DMMS.R – The Epsilon Group’. Accessed: Nov. 16, 2023. [Online]. Available: https://tegvirginia.com/software/dmms-r/

[29] J. Xie, *Simglucose v0.2.1*. (2018). Python. Accessed: Jan. 16, 2025. [Online]. Available: https://github.com/jxx123/simglucose

[30] T. Zhu, K. Li, P. Herrero, and P. Georgiou, ‘Basal Glucose Control in Type 1 Diabetes Using Deep Reinforcement Learning: An In Silico Validation’, *IEEE J Biomed Health Inform*, vol. 25, no. 4, pp. 1223–1232, Apr. 2021, doi: 10.1109/JBHI.2020.3014556.

[31] R. Hovorka *et al.*, ‘Partitioning glucose distribution/transport, disposal, and endogenous production during IVGTT’, *Am J Physiol Endocrinol Metab*, vol. 282, no. 5, pp. E992-1007, May 2002, doi: 10.1152/ajpendo.00304.2001.

[32] ‘A Model-Free Approach to Automatic Dose Guidance in Long Acting Insulin Treatment of Type 2 Diabetes | IEEE Journals & Magazine | IEEE Xplore’. Accessed: Jan. 16, 2025. [Online]. Available: https://ieeexplore.ieee.org/document/9308924

[33] ‘A Deep Reinforcement Learning Approach for Type 2 Diabetes Mellitus Treatment | IEEE Conference Publication | IEEE Xplore’. Accessed: Jan. 16, 2025. [Online]. Available: https://ieeexplore.ieee.org/document/9374313

[34] F. Lv *et al.*, ‘Characteristics of Newly Diagnosed Type 2 Diabetes in Chinese Older Adults: A National Prospective Cohort Study’, *J Diabetes Res*, vol. 2019, p. 5631620, 2019, doi: 10.1155/2019/5631620.

[35] H. Emerson, M. Guy, and R. McConville, ‘Offline reinforcement learning for safer blood glucose control in people with type 1 diabetes’, *Journal of Biomedical Informatics*, vol. 142, p. 104376, Jun. 2023, doi: 10.1016/j.jbi.2023.104376.

[36] S. Del Giorno, F. D’Antoni, V. Piemonte, and M. Merone, ‘A New Glycemic closed-loop control based on Dyna-Q for Type-1-Diabetes’, *Biomedical Signal Processing and Control*, vol. 81, p. 104492, Mar. 2023, doi: 10.1016/j.bspc.2022.104492.

[37] M. Panagiotou *et al.*, ‘A Complete AI-Based System for Dietary Assessment and Personalized Insulin Adjustment in Type 1 Diabetes Self-management’, 2023, pp. 77–86. doi: 10.1007/978-3-031-44240-7_8.

[38] A. Jafar, M.-R. Pasqua, B. Olson, and A. Haidar, ‘Advanced decision support system for individuals with diabetes on multiple daily injections therapy using reinforcement learning and nearest-neighbors: In-silico and clinical results’, *Artif Intell Med*, vol. 148, p. 102749, Feb. 2024, doi: 10.1016/j.artmed.2023.102749.

[39] J. Yoo, V. Pradana Rachim, S. Jung, and S.-M. Park, ‘Intelligent Dual Basal–Bolus Calculator for Multiple Daily Insulin Injections via Offline Reinforcement Learning’, *IEEE Access*, vol. 12, pp. 192572–192585, 2024, doi: 10.1109/ACCESS.2024.3518832.

[40] Q. Sun, M. V. Jankovic, and S. G. Mougiakakou, ‘Reinforcement Learning-Based Adaptive Insulin Advisor for Individuals with Type 1 Diabetes Patients under Multiple Daily Injections Therapy’, *Annu Int Conf IEEE Eng Med Biol Soc*, vol. 2019, pp. 3609–3612, Jul. 2019, doi: 10.1109/EMBC.2019.8857178.

[41] A. Jafar, A. Kobayati, M. A. Tsoukas, and A. Haidar, ‘Personalized insulin dosing using reinforcement learning for high-fat meals and aerobic exercises in type 1 diabetes: a proof-of-concept trial’, *Nat Commun*, vol. 15, no. 1, p. 6585, Aug. 2024, doi: 10.1038/s41467-024-50764-5.

[42] P. Avari *et al.*, ‘Safety and Feasibility of the PEPPER Adaptive Bolus Advisor and Safety System: A Randomized Control Study’, *Diabetes Technol Ther*, vol. 23, no. 3, pp. 175–186, Mar. 2021, doi: 10.1089/dia.2020.0301.

[43] P. Herrero, P. Pesl, M. Reddy, N. Oliver, P. Georgiou, and C. Toumazou, ‘Advanced Insulin Bolus Advisor Based on Run-To-Run Control and Case-Based Reasoning’, *IEEE J Biomed Health Inform*, vol. 19, no. 3, pp. 1087–1096, May 2015, doi: 10.1109/JBHI.2014.2331896.

[44] L. Eiland, M. McLarney, T. Thangavelu, and A. Drincic, ‘App-Based Insulin Calculators: Current and Future State’, *Curr Diab Rep*, vol. 18, no. 11, p. 123, Oct. 2018, doi: 10.1007/s11892-018-1097-y.